\pdfoutput=1
\PassOptionsToPackage{unicode}{hyperref}
\PassOptionsToPackage{hyphens}{url}
\PassOptionsToPackage{dvipsnames,svgnames,x11names}{xcolor}
\documentclass[
  11pt,
]{article}

\usepackage{amsmath,amssymb}
\usepackage{iftex}
\ifPDFTeX
  \usepackage[T1]{fontenc}
  \usepackage[utf8]{inputenc}
  \usepackage{textcomp} 
\else 
  \usepackage{unicode-math}
  \defaultfontfeatures{Scale=MatchLowercase}
  \defaultfontfeatures[\rmfamily]{Ligatures=TeX,Scale=1}
\fi
\usepackage{lmodern}
\ifPDFTeX\else  
\fi
\IfFileExists{upquote.sty}{\usepackage{upquote}}{}
\IfFileExists{microtype.sty}{
  \usepackage[]{microtype}
  \UseMicrotypeSet[protrusion]{basicmath} 
}{}
\usepackage{xcolor}
\makeatletter
\ifx\paragraph\undefined\else
  \let\oldparagraph\paragraph
  \renewcommand{\paragraph}{
    \@ifstar
      \xxxParagraphStar
      \xxxParagraphNoStar
  }
  \newcommand{\xxxParagraphStar}[1]{\oldparagraph*{#1}\mbox{}}
  \newcommand{\xxxParagraphNoStar}[1]{\oldparagraph{#1}\mbox{}}
\fi
\ifx\subparagraph\undefined\else
  \let\oldsubparagraph\subparagraph
  \renewcommand{\subparagraph}{
    \@ifstar
      \xxxSubParagraphStar
      \xxxSubParagraphNoStar
  }
  \newcommand{\xxxSubParagraphStar}[1]{\oldsubparagraph*{#1}\mbox{}}
  \newcommand{\xxxSubParagraphNoStar}[1]{\oldsubparagraph{#1}\mbox{}}
\fi
\makeatother

\providecommand{\tightlist}{%
  \setlength{\itemsep}{0pt}\setlength{\parskip}{0pt}}\usepackage{longtable,booktabs,array}
\usepackage{calc} 
\usepackage{etoolbox}
\makeatletter
\patchcmd\longtable{\par}{\if@noskipsec\mbox{}\fi\par}{}{}
\makeatother
\IfFileExists{footnotehyper.sty}{\usepackage{footnotehyper}}{\usepackage{footnote}}
\makesavenoteenv{longtable}
\usepackage{graphicx}
\makeatletter
\def\maxwidth{\ifdim\Gin@nat@width>\linewidth\linewidth\else\Gin@nat@width\fi}
\def\maxheight{\ifdim\Gin@nat@height>\textheight\textheight\else\Gin@nat@height\fi}
\makeatother
\setkeys{Gin}{width=\maxwidth,height=\maxheight,keepaspectratio}
\makeatletter
\def\fps@figure{htbp}
\makeatother

\makeatletter
\@ifpackageloaded{caption}{}{\usepackage{caption}}
\AtBeginDocument{%
\ifdefined\contentsname
  \renewcommand*\contentsname{Table of contents}
\else
  \newcommand\contentsname{Table of contents}
\fi
\ifdefined\listfigurename
  \renewcommand*\listfigurename{List of Figures}
\else
  \newcommand\listfigurename{List of Figures}
\fi
\ifdefined\listtablename
  \renewcommand*\listtablename{List of Tables}
\else
  \newcommand\listtablename{List of Tables}
\fi
\ifdefined\figurename
  \renewcommand*\figurename{Figure}
\else
  \newcommand\figurename{Figure}
\fi
\ifdefined\tablename
  \renewcommand*\tablename{Table}
\else
  \newcommand\tablename{Table}
\fi
}
\@ifpackageloaded{float}{}{\usepackage{float}}
\floatstyle{ruled}
\@ifundefined{c@chapter}{\newfloat{codelisting}{h}{lop}}{\newfloat{codelisting}{h}{lop}[chapter]}
\floatname{codelisting}{Listing}

\makeatother
\makeatletter
\@ifpackageloaded{caption}{}{\usepackage{caption}}
\@ifpackageloaded{subcaption}{}{\usepackage{subcaption}}
\makeatother
\makeatletter
\@ifpackageloaded{algorithm}{}{\usepackage{algorithm}}
\makeatother
\makeatletter
\@ifpackageloaded{algpseudocode}{}{\usepackage{algpseudocode}}
\makeatother
\ifLuaTeX
  \usepackage{selnolig}  
\fi
\usepackage[]{natbib}
\usepackage{bookmark}

\IfFileExists{xurl.sty}{\usepackage{xurl}}{} 
\hypersetup{
  pdftitle={HSAP: A Hierarchical Sequence-aware Parallelism for Hybrid-Context Generative Models},
  pdfkeywords={Machine Learning, EMNLP},
  colorlinks=true,
  linkcolor={blue},
  filecolor={Maroon},
  citecolor={Blue},
  urlcolor={Blue},
  pdfcreator={LaTeX via pandoc}}

\usepackage{times}
\usepackage{latexsym}

\usepackage{inconsolata}
\usepackage{multirow}
\usepackage[preprint]{acl}

\title{HSAP: A Hierarchical Sequence-aware Parallelism for Hybrid-Context
Generative Models}
\author{
  \textbf{Songxin Zhang \textsuperscript{1}\thanks{Equal Contribution}},
  \textbf{Zejian Xie \textsuperscript{1}\footnotemark[1]\thanks{Correspending Author}},
  \textbf{Zhuoyang Song\textsuperscript{1}\footnotemark[1]},
\\
  \textbf{Cong lin\textsuperscript{1}},
  \textbf{Junyu Lu\textsuperscript{1,2}},
  \textbf{Jiaxing Zhang\textsuperscript{2}},
  \textbf{Bingyi Jing \textsuperscript{1}\footnotemark[2]},
\\
\\
  \textsuperscript{1}Department of Statistics and Data Science, Southern University of Science and Technology
\\
  \textsuperscript{2}International Digital Economy Academy
\\
  \small{
    \textbf{Correspondence:} \href{mailto:xiezj@mail.sustech.edu.cn}{xiezj@mail.sustech.edu.cn} \href{mailto:jingby@sustech.edu.cn}{jingby@sustech.edu.cn}
  }
}

\date{}

\begin{document}
\maketitle
\begin{abstract}
In this paper, we aim to combine the advantages of existing sequence
parallelism paradigms and overcomes their drawbacks, the most serious of
which is the incapability to correctly compute causal attention on the
hybrid-context packed sequences, in a stronger sequence parallelism
framework. The practical technique of packing sequences for efficiently
pretraining and fine-tuning large language models causes
cross-contamination problem in attention computation, which can be
effectively solved when no parallelism in the sequence length dimension
is taken. However, in sequence parallelism, existing approaches either
ignore the scenario of hybrid-context sequences or conversely sacrifice
and limit parallelism degree for supporting the scenario. To this end,
we innovatively propose an efficient Sequence-Aware Parallelism
algorithm to conquer the obstacles of intensive tensor transmission and
partial attention computation across multiple device groups. Our
algorithm utilizes JIT (Just-In-Time) compilation to optimize the
communication strategy of all device groups in NCCL level. Further, we
integrate existing sequence parallelism paradigms into a Hierarchical
Sequence-Aware Parallelism framework which benefits from our
sequence-aware algorithm. We additionally elaborate on the memory and
communication overhead management of the hierarchical framework to
optimize its performance. Through multiple experiments, we demonstrate
that our proposed approach outperform other state-of-the-arts sequence
parallelism approches in multiple metrics.
\end{abstract}

\floatname{algorithm}{Algorithm}

\section{Introducion}\label{introducion}

Numerous complex generative tasks have necessitated modelling over long
context in both spatial and temporal domains, driving the trend of
generative models capable of handling long sequences, particularly in
multi-modal foundational models that process speech, images, and
wavelengths \citep{liu2023ls_ex1, li2023ls_ex2, xiong2023ls_ex3}. Large
language models (LLMs) like Claude and
ChatGPT\citep{6G5Z7P8U_Openai_Etal_2023, brown2020gpt3} have
pioneeringly extended sequence length to over 100K tokens, while
multi-modal models Gemini\citep{4ZW46HK7_GeminiTeam_Etal_2023m} and
Sora\citep{videoworldsimulators2024} have propelled this trend to exceed
1M tokens. However, training on ultra-long sequences presents a
challenge for existing distributed systems due to the significant
increase in activation memory footprints \citep{liu2024world}.

Recently, milestone sequence parallelism (SP) works such as
Deepspeed-Ulysess \citep{jacobs2023deepspeed} and Ring-Attention
\citep{liu2023ring_attn} have been widely adopted, alleviating memory
stress by partitioning activations along the sequence length dimension.
They have standardized SP into two paradigms with Deepspeed-Ulysess
using all-to-all communication to compute attention within the parallel
group (intra-group) and Ring-Attention using peer-to-peer (P2P)
communication to compute attention across parallel groups (inter-group).
However, the parallelism degree of DeepSpeed-Ulysses is up-bounded by
the number of attention heads, and the predefined topology in
Ring-Attention increases the communication and computation overhead.
More importantly, the common practice of packing unrelated
hybrid-context sequences into longer sequencesm, which saves the
computational waste on padded tokens in batch training, does not
function properly with SP due to the cross-contamination attention
problem.

In this paper, we address the challenges of sequence parallelism on
hybrid-context long packed sequences for training LLMs. We first present
a comprehensive analysis on sequence packing and the existing SP
paradigms, and point out that we can combine the complimentary
advantages of those paradigms in a novel SP framework and adapt it to
the correct attention computation on hybrid-context packed sequences. To
this end, we invent an innovative \textbf{S}equence-\textbf{A}ware
\textbf{P}arallelism (SAP) algorithm that utilizes \textbf{JIT}
(just-in-time) compilation idea to realize the extremely complicated
scheduling and organization of incomplete QKV tensors transmission and
irregular attention patterns computation. Then, we hierarchically combine
the existing SP paradigms in the \textbf{H}ierarchical
\textbf{S}equence-\textbf{A}ware \textbf{P}arallelism (HSAP) framework
to obtain a stronger paradigm as well as benefits from the SAP. Our main
contributions are summarized as below:

\begin{itemize}
\tightlist
\item
  We propose the SAP algorithm, a fast and efficient algorithm to
  compute exact attention on hybrid-context packed sequences
  distributedly. SAP follows the idea of JIT compilation to generate a
  series of operation codes for each distributed rank so that the
  sophisticated P2P communications can be arranged efficiently and
  harmonically.
\item
  We develop the HSAP framework which integrates the benefits of both
  inter-group and intra-group sequence parallelism. The framework
  combines SAP mechanism with improved memory and communication
  management and provide a scalable solution for handling extremely long
  sequences for training LLMs.
\item
  We conduct a number of informative experiments to manifest the
  effectiveness of our proposed SAP algorithm and HSAP framework.
  Performance results show that our approach can extend sequence length
  to over 512K, outperforming other state-of-the-art sequence
  parallelism approaches. We achieve the highest efficiency in terms of
  TGS (tokens per GPU second) when training on real hybrid-context
  datasets compared with vanilla sequence packing and batch training. We
  also achieve higher speed of attention computation on packed sequences
  than the existing SP paradigms.
\end{itemize}

\section{Preliminaries}\label{preliminaries}

We begin with the ubiquitous practice of sequence packing, which serves
as the motivation for our sequence-aware parallelism. Then, we introduce
two different designs that mark the milestone in sequence parallelism
development.

\subsection{Sequence Packing}\label{sequence-packing}

To ensure efficiency of training on length-variable sequences, the
practice of sequence packing is commonly adopted in both pretraining and
fine-turning cases by concatenating multiple short sequences to fill the
max sequence length as much as possible {[}\citet{Raffel_2019};
\citet{MBSTSUBQ_Krell_Etal_2022}; \citet{78WUPJSC_Shi_Etal_2023};
Staniszewski\_2023{]}. Besides efficiency, this approach has multiple
significant advantages over the alternative batching approach where
random or grouped-by-length sequences are padded to meet the max
sequence length in the batch \citep{MBSTSUBQ_Krell_Etal_2022}. For
example, by retrieving and packing related documents to form
hybrid-context sequences for pretraining, LLMs can catch longer context
dependency \citep{78WUPJSC_Shi_Etal_2023, Staniszewski_2023}.

\begin{figure}

\centering{

\includegraphics{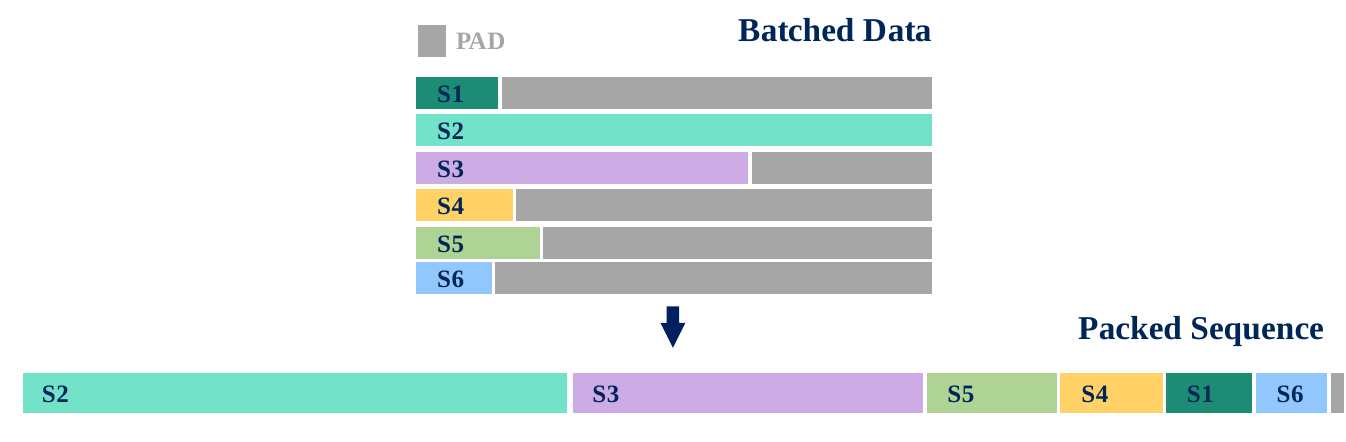}

}

\caption{\label{fig-packing}Packing hybrid-context data into single
sequence reduces computational waste caused by padding.}

\end{figure}%

However, in general cases, the use of sequence packing is hindered by
the cross-contamination attention problem that the hybrid-context
segments in the packed sequence attend to each other unexpectedly. As
shown in Figure~\ref{fig-packing}, to match the behavior of causal
attention, it requires custom kernel design or unpacking the sequence
back to batch \citep{MBSTSUBQ_Krell_Etal_2022}, which is either
complicated or inefficient. Thanks to FlashAttention
\citep{dao2022flashattention}, the challenge can be effectively dealed
with using \emph{cu\_len} (cumulative length) methods to only compute
attention within each segment as long as the packed sequence can fit
into a single GPU. Unfortunately, with recent advances in LLMs and
multi-modal models, the context length has unprecedentedly grown so long
that even without packing, a single sequence may not fit into a GPU.
Parallelism along the sequence length dimension is hence necessitated
and it confronts sequence packing with new issues.

\subsection{Sequence Parallelism}\label{sequence-parallelism}

Sequence parallelism has been proposed to slice long sequences, scatter
the segments to multiple device groups, and parallelize attention
computation. Its challenge lies in the trade-off between lowering memory
cost for activation and maintaining communication cost. Early attempts
feature Megatron-LM \citep{korthikanti2023megatronlm} who implements
sequence parallelism as an extension to its tensor parallelism, an
unsatisfactory way that the high complexity attention is not
parallelized in the length dimension. But by now, sequence parallelism
has evolved into two distinct paradigms which we summarize as
Inter-group sequence parallelism (InterSP) and Intra-group sequence
parallelism (IntraSP).

\subsubsection{Inter-group Sequence
Parallelism}\label{inter-group-sequence-parallelism}

Recall attention calculation \citep{vaswani2017transformer}. Given
query, key, and value sequences
\(Q, K, V \in \mathbb{R}^{N\times H\times d}\) where \(N\), \(H\) and
\(d\) represent the length, the number of attention heads and the head
size respectively. In single-head case with \(H = 1\), the attention is
given by
\[\text{Attention}(Q, K, V) = \text{Softmax}\left(\frac{QK^T}{\sqrt{d}}\right)V,\]
while in multi-head case, \(H\) single-head attentions are first
computed respectively in the number of heads dimension and the outputs
are then concatenated along the head size dimension followed by a linear
projection.

In InterSP, the sliced blocks
\(Q_i, K_i, V_i \in \mathbb{R}^{n \times H \times d}\) where
\(n = N / P_\text{inter}\) and \(i \leq P_\text{inter}\) are allocated
to \(P_\text{inter}\) device groups permanently without reunifying the
complete sequences during computation. For the locally stored \(Q_i\),
\(K_j\) and \(V_j\) blocks in other groups are still needed in
attention, and InterSP approaches
\citep{li-etal-2023-seq_parallel, liu2023ring_attn, li2023colossalai, li2023lightseq}
leverage P2P communucation in a ring-style topology with each rank
sending and receiving \(K_j\) and \(V_j\) and computing attention
concurrently in an online manner \citep{Milakov_Gimelshein_2018}.

While InterSP achieves maximal efficiency by overlapping communication
and computation process, its ring-style topology is incompatible with
sequence packing because the predefined topology determines that only in
the case where the local segment should attend to all previous segments
can it function properly. For a long packed sequence, segments with
unrelated hybrid context should not attend to each other, and InterSP
will either take unnecessary overhead to compute incorrect attention, or
integrate intractable strategies to skip some communications.

\subsubsection{Intra-group Sequence
Parallelism}\label{intra-group-sequence-parallelism}

Compared to InterSP, IntraSP maintains the entire sequence in each group
for attention computation but instead distributes the heads across
groups. Formally, before computing attention, the sliced blocks
\(Q_i, K_i, V_i \in \mathbb{R}^{n \times H \times d}\) with
\(n = N / P_\text{intra}\) and \(i \leq P_\text{intra}\) are maintained
among \(P_\text{intra}\) device groups. IntraSP (e.g.
\citep{jacobs2023deepspeed}) then leverages all-to-all operation to move
the parallelism from the length \(n\) to the number of heads \(H\) and
obtain blocks
\(Q^{\text{intra}}_i, K^{\text{intra}}_i, V^{\text{intra}}_i \in \mathbb{R}^{N \times h \times d}\)
with \(h = H / P_\text{intra}\). Since each block now has the complete
length \(N\), multi-head attention can be directly computed within the
local rank. After that, another all-to-all operation moves the
parallelism back to the length dimension as it is in the initial blocks.

Unlike InterSP, IntraSP has no problems with packed sequences because it
can use FlashAttention as usual to compute multi-head attention for the
complete sequence within a single group. However, it is obvious that its
parallelism degree \(P_\text{intra}\) cannot exceed the number of heads
\(H\), which seriously limits its practicality as \(N\) is usually much
larger than \(H\).

\section{Adapt Sequence Parallelism to Hybrid
Context}\label{adapt-sequence-parallelism-to-hybrid-context}

We aim to build a new paradigm for sequence parallelism that combines
InterSP and IntraSP and overcomes their drawbacks. To this end, we first
propose the novel SAP algorithm to make InterSP function properly with
packed sequences. Then, we present our combined sequence parallelism in
the HSAP framework with SAP mechanism and improved management on memory
and communication.

\subsection{Sequence-aware Parallelism via Just-in-time
Compilation}\label{sequence-aware-parallelism-via-just-in-time-compilation}

The primary challenge is that the ring-style attention on packed
sequences involves diverse attention patterns and transmission of
incomplete \(Q_i, K_i, V_i\) across ranks. It will be sophisticated and
not robust to program the transmission scheduling and computation by
inspecting the miscellaneous conditional branches. Inspired by JIT
compilation, we instead propose to compile a series of codes in a
sequence-aware way ahead of the communitation and computation.

\begin{figure*}

\centering{

\includegraphics{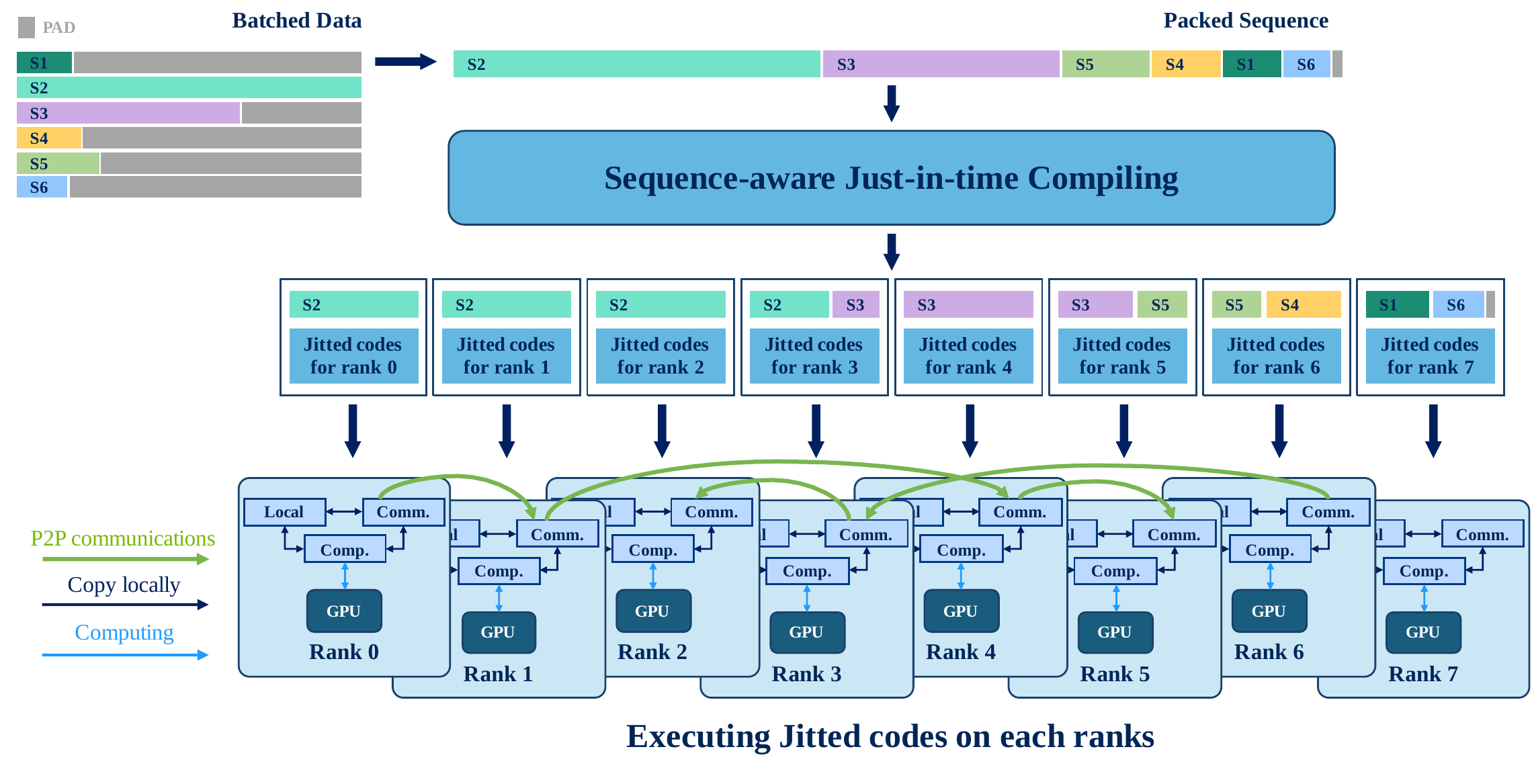}

}

\caption{\label{fig-jitframework}SAP's just-in-time compile-execute
architecture. According to the structure of hybrid-context, attention is
compiled just-in-time into distributed computation and communication
operations, forming jitted codes.Each rank performs peer-to-peer
communication, computation, and memory operations according to its own
codes.}

\end{figure*}%

Our JIT compile-execute architecture for SAP is shown in
Figure~\ref{fig-jitframework}. Given a packed sequence, our SAP
algorithm dynamically generate a code series for each rank respective.
When executing the series step by step, all ranks cooperate in harmony.
By fusing operations and reducing bubbles, the compilation can reduce
the communication cost of all ranks in NCCL level.

\begin{table}

\centering{

\centering
\resizebox{\columnwidth}{!}{
\begin{tabular}{|l|l|}
\hline
\multicolumn{2}{|c|}{\textbf{Communication}} \\ \hline
\textbf{Syntax} & create \$send-op \$\{q, kv, o, dq, dkv, do\} \%\{dst\} \\ \hline
\textbf{Semantics} & Create a peer-to-peer operation, sending \{q, kv, o, dq, dkv, or do\} to \{dst\} rank. \\ \hline
\textbf{Syntax} & create \$recv-op \$\{q, kv, o, dq, dkv, do\} \%\{src\} \\ \hline
\textbf{Semantics} & Create a peer-to-peer operation, receiving \{q, kv, o, dq, dkv, or do\} from \{src\} rank. \\ \hline
\multicolumn{2}{|c|}{\textbf{Computing}} \\ \hline
\textbf{Syntax} & computing \\ \hline
\textbf{Semantics} & Perform attention computation using tensors within the computing buffer. \\ \hline
\multicolumn{2}{|c|}{\textbf{Copy locally}} \\ \hline
\textbf{Syntax} & copy-\${q, kv, o, dq, dkv, do} \$\{buffer\_src\} \$\{buffer\_dst\} \\ \hline
\textbf{Semantics} & Locally copy \{q, kv, o, dq, dkv, do\} from the src buffer to the dst buffer. \\ \hline
\end{tabular}
}

}

\caption{\label{tbl-instructions}Definition of instructions used in jitted codes.}

\end{table}%

Concretely, SAP can catch a src rank's need to P2P transfer KV tensors
to a dst rank. At the same position in the series of src and dst rank,
SAP places \emph{create-\$\{send\}-op \$\{kv\} \%\{dst\}} code and
\emph{create-\$\{receive\}-op \$\{kv\} \%\{src\}} code respectively.
When computing attention, \emph{copy-\$\{kv\}-\$\{comm\}-\$\{comp\}}
moves KV tensors between buffers, and \emph{compute} orders the rank to
compute flash attention on QKV tensors in the buffer. More details in
Table~\ref{tbl-instructions}.

\subsection{Compiling for Computing-efficient Communication
Strategies}\label{compiling-for-computing-efficient-communication-strategies}

\begin{figure*}

\centering{

\includegraphics{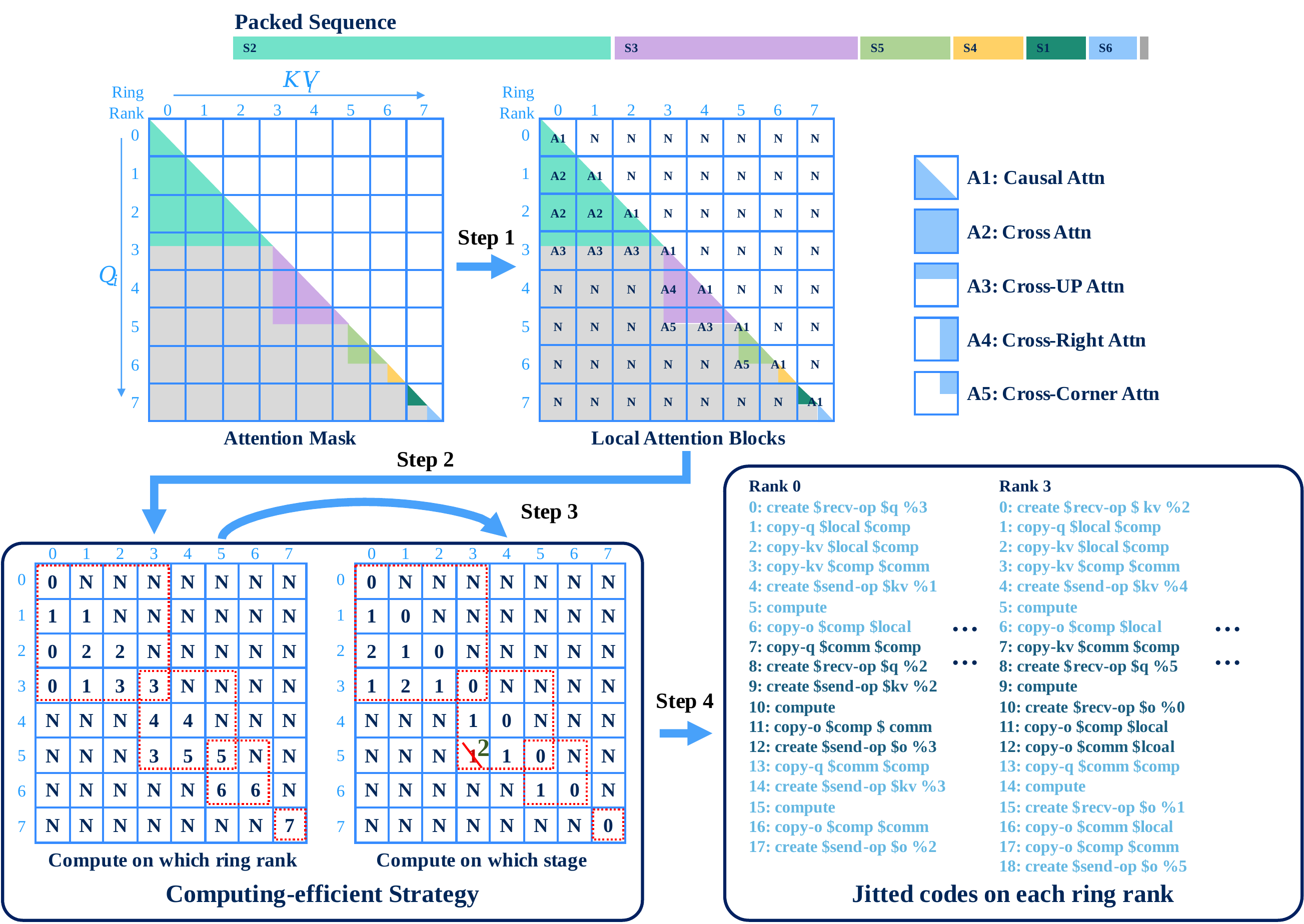}

}

\caption{\label{fig-jitcompiling}Compilation Algorithms for
Computationally Efficient Communication Strategies.}

\end{figure*}

Based on our SAP JIT compiling architecture, the efficiency improvement
of sequence parallelism in hybrid-context scenarios comes from the
quality of the jitted code, which needs to consider the specific
structure of the packed sequence and trim redundant communication and
computation. Shown in Figure~\ref{fig-jitcompiling}, we propose a
compiling algorithm for computation-friendly communication strategy that
takes the sequence topology, sequence parallel world size, max sequence
length per rank as inputs to generate efficient execution instructions
for each rank.

\textbf{Partition attention computation blocks.} Consider the complete
attention mask that is split across InterSP ranks. Naturally, we
partition the attention computation into local attention block based on
the distribution of sequences across different InterSP ranks. As is
shown in Figure~\ref{fig-jitcompiling} Step 1, we can infer from the map
that there are only 5 kinds of attention patterns for each rank block in
the map (i.e.~A1\textasciitilde A5). Based on this, we can apply
different computation strategies in different ranks.

\textbf{Generate Regional Strategy and Resolve Global Conflict.} Based
on the partitioned attention blocks, we assign the computation tasks of
each attention block to SAP ranks according to the sequence and
ring-attention load balancing.

As shown in the bottom-left diagram of Figure~\ref{fig-jitcompiling},
step 2 conduct workload balancing within each causal attention (red
dashed box in Figure~\ref{fig-jitcompiling}), where the specific load
balancing strategy is similar to the scheduling algorithm in
\citep{li2023lightseq} With this arrangement, we approximately achieve
load balancing of the computation overhead (the number of attention
blocks to be computed in the figure) for each rank. Step 2 resulted in a
layout showing which rank computes each attention block.

Globally, regional load balancing might cause conflicts for ranks in
overlapping regions (i.e.~rank 3 in Figure~\ref{fig-jitcompiling}). Step
3 address conflicts caused by an SAP rank being required to two reagion
(i.e.~region 0\textasciitilde3 and region 3\textasciitilde5), while also
striving to balance the computation load across all SAP ranks globally.
Ultimately, step 3 resulted in a layout showing the computation order of
attention blocks (a.k.a. computing on which stage).

\textbf{Jitted code Generation.} The objective of this step is to
compile the computation layout of attention blocks obtained in the
previous step into the final operation sequence list that needs to be
passed to the HSAP attention operator. These lists correspond to the
instruction set of Table~\ref{tbl-instructions}, as shown in the lower
right corner of Figure~\ref{fig-jitcompiling}.

In this phrase, we arrange the communication and computation tasks for
each computation stage based on the tasks of current computation stage
and the next stage. Specifically, we aim to have a legitimate
communication task occurring simultaneously with the computation task
whenever possible. This includes the transmission of computation results
(O/dQ/dKV) or data (Q, KV, dO, etc.) from other SAP Ranks for the next
stage. The initiation of this communication task requires the
send/receive pair to be activated simultaneously between the two SAP
ranks involved in the P2P communications. When handling batched P2P
operations within a group, NCCL coalesces them into efficient group
operations to maximize the utilization of NVLink bandwidth. Therefore,
when compiling into the actual intruction series, we strive to ensure
that each \emph{batch\_isend\_irecv} operation covers the entire SAP
group without conflicts.

\subsection{Hierarchically Combine InterSP and
IntraSP}\label{hierarchically-combine-intersp-and-intrasp}

\begin{figure}

\centering{

\includegraphics{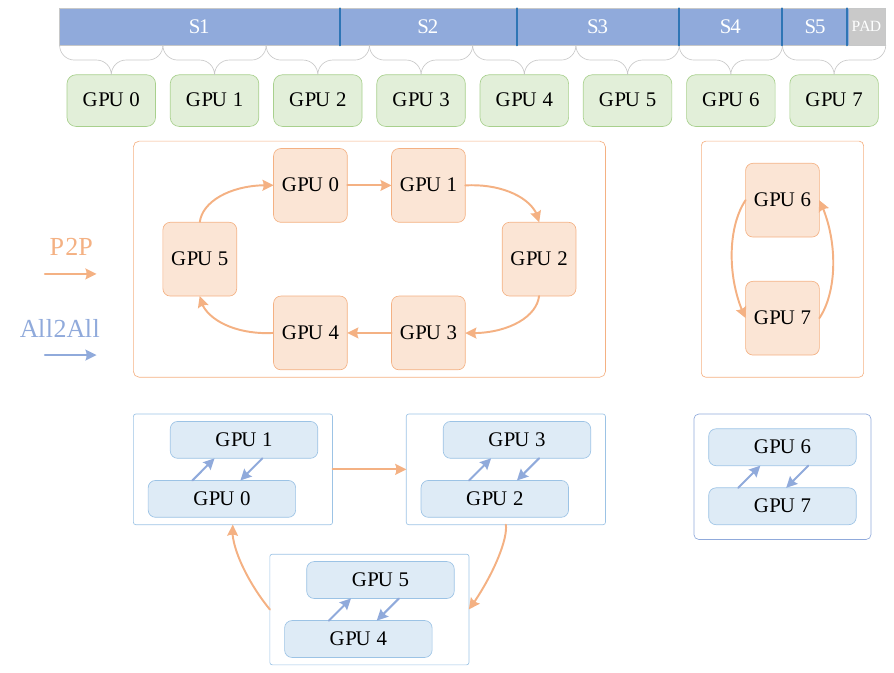}

}

\caption{\label{fig-ljy}The hierarchical network hardware topology.}

\end{figure}%

As shown in Figure~\ref{fig-ljy}, we now present HSAP to integrate
IntraSP into InterSP in a hierarchical way for pursuing higher sequence
parallelism degree and benefits from SAP. In HSAP, we coordinate InterSP
and IntraSP to partition sequences along the length dimension. The total
SP groups are now composed of two orthogonal groups: InterSP at the
upper level and IntraSP at the lower level.

The \(Q, K, V\) sequences are first distributed across
\(P_\text{inter}\) InterSP groups, each of which maintains
\(Q^{\text{inter}}_i, K^{\text{inter}}_i, V^{\text{inter}}_i \in \mathbb{R}^{(N/P_\text{inter}) \times H \times d}\)
and utilizes P2P for necessary block transmission. Each group is further
divided into \(P_\text{intra}\) IntraSP groups that respectively
maintain
\(Q^{\text{inter}}_{i,j}, K^{\text{inter}}_{i,j}, V^{\text{inter}}_{i,j} \in \mathbb{R}^{n \times H \times d}\),
where \(n = N/(P_\text{inter}\times P_\text{intra})\) and
\(i \leq P_\text{inter}, j \leq P_\text{intra}\) are indices in the
length dimension. The actual computation happens in this level with an
all-to-all operation collecting and reforming
\(Q^{\text{inter}}_{i,j}, K^{\text{inter}}_{i,j}, V^{\text{inter}}_{i,j}\)
into
\(Q^{\text{intra}}_{i,j}, K^{\text{intra}}_{i,j}, V^{\text{intra}}_{i,j} \in \mathbb{R}^{(N/P_\text{inter}) \times (H/P_\text{intra}) \times d}\),
where \(j\) becomes index in the number of heads dimension. Then, we can
utilize the InterSP level P2P for computing
\(\text{Attention}(Q^{\text{intra}}_j, K^{\text{intra}}_j, V^{\text{intra}}_j)\)
where the arguments have the shape
\(N \times (H/P_\text{intra}) \times d\). Since this attention has
nothing to do across IntraSP groups, SAP can be smoothly adopted.

\subsection{Manage Memory and
Communication}\label{manage-memory-and-communication}

To further control memeory peak and accelerate communication, we design
a novel method in HSAP for memory and communication management. We first
follow \citep{li2023lightseq} to balance workload of different InterSP
groups by additionally transfering \(Q_i\) blocks. Upon that, in each
group, we allocate a global memory buffer for storing the local
\(Q_i, K_i, V_i\) and the received \(Q_j, K_j, V_j\). The size of the
buffer is a fixed multiple of \((N/P_\text{inter}) \times H \times d\)
such that there is no memory jitter during transmissions. With
\(N / P_\text{inter}\) fixed, HSAP can constantly extend sequence length
with controllable memory peak. ~

To optimize the hierarchical communication effecacy of HSAP, we
distribute InterSP and IntraSP groups on clusters with different network
infrastructures based on the properties of P2P and all-to-all operators.
IntraSP groups with communication-intensive all-to-all are placed among
closely situated ranks connected via NVLink, and InterSP groups with
communication-moderate P2P are placed among ranks with lower bandwidth
connected via InfiniBand.

\section{Experiments}\label{experiments}

In this section, we evaluate the performance of our proposed SAP
algorithm together with the HSAP framework from the perspectives of the
performance on hybrid context, attention computation efficiency, and
scalability to ultra-long sequences on large LLM.

\textbf{Baselines.} Our experiments focus on comparisons with recent
state-of-the-art sequence parallel algorithms and systems listed as
follows:

\begin{itemize}
\tightlist
\item
  We compare with DeepSpeed-Ulysses \citep{jacobs2023deepspeed}, the
  pioneer and dominator of IntraSP which is integrated into the
  DeepSpeed system \citep{Rasley_Rajbhandari_Ruwase_He_2020} for
  parallelizing large model training.
\item
  We compare with the DistFlashAttn \citep{li2023lightseq}, the latest
  InterSP algorithm that distributes flash attention and improves the
  load scheduling of the ring-style communication topology.
\item
  We also compare with ColAI \citep{li2023colossalai} and Megatron-LM
  \citep{shoeybi2019megatronlm, korthikanti2023megatronlm},
  comprehensive distributed training systems which support InterSP with
  the implementation of ring-attention \citep{liu2023ring_attn} and also
  other parallelism including tensor and pipeline parallelism. Note that
  sequence parallelism is orthogonal to other parallelisms which are
  configured optimally if needed.
\end{itemize}

\textbf{Model setup.} We evaluete the proposed HSAP framework on a 30B
GPT model in line with other systems for sequence length scalability
comparison. For other experiments, we use LLaMA2 and its variants which
are more common architectures in practice.

\textbf{Hardware setup.} Depending on the resources required for each
experiment, we conduct experiments on the following two hardware
platforms:

\begin{enumerate}
\def\labelenumi{\arabic{enumi}.}
\tightlist
\item
  Eight DGX nodes, each with 8 x A100 40G GPUs. GPUs within the nodes
  are connected via NVLink, and nodes are connected via Infiniband.
\item
  A single HGX node equipped with 8 x A100 80G GPUs. GPUs within the
  nodes are connected via NVLink.
\end{enumerate}

\subsection{Attention Performance
Evaluation}\label{attention-performance-evaluation}

We first experimented with the efficiency improvement of SAP in
hybrid-context scenarios. We conduct an experiment on 8×A100 devices to
compare the single attention computation time between DisFlashAttn and
sequence-aware parallelism (SAP). Empirically, we set the
\emph{num\_head} to 16 and \emph{hidden\_size} to 128. For a fair
comparison, we take the maximum attention computation time across all
ranks and calculate the average time (ms) for 1000 forward and backward
passes as the experiment result.

\begin{table}

\centering{

\centering
\resizebox{\columnwidth}{!}{%
\begin{tabular}{@{}ccccccccc@{}}
\toprule
\textbf{Max Seq Len} & \textbf{8K}  & \multicolumn{2}{c}{\textbf{16K}} & \multicolumn{2}{c}{\textbf{32K}} & \multicolumn{2}{c}{\textbf{64k}} & \textbf{128K}   \\ \midrule
\textbf{SP size}       & \textbf{2} & \textbf{2} & \textbf{4} & \textbf{4} & \textbf{8} & \textbf{8} & \textbf{16} & \textbf{16} \\ \midrule
\textbf{DistFlashAttn} & 9.0        & 24.5       & 16.2       & 36.2       & 112.0      & 218.1      & 4373.7      & 8254.5      \\
\textbf{Ours}        & \textbf{7.9} & \textbf{14.6}   & \textbf{11.1}  & \textbf{25.8}   & \textbf{73.0}  & \textbf{137.3} & \textbf{1912.3} & \textbf{3333.1} \\ \bottomrule
\end{tabular}
}

}

\caption{\label{tbl-hybridsingleattn}Computation efficiency of single attention in different Hybrid-context scenarios, measured by milliseconds.}

\end{table}%

As illustrated in Table~\ref{tbl-hybridsingleattn}, We can observe that
as SP size increases from 2 to 16 and the maximum sequence length
increases from 8K to 256K, the computation time of both SAP and
DisFlashAttn exhibits near-exponential growth due to the increased
communication and computation load. Owing to sequence-aware just-in-time
compilation, SAP can coordinate all ranks during communication and
computation to fuse operations and reduce bubbles, significantly
mitigating the exponential growth rate of attention computation.

Then, we also conducted experiments on the efficiency of HSAP at the
granularity of a single attention, comparing it with Ulysses and
DistFlashAttn. Specifically, we scale up the sequence length per GPU,
the total number of GPUs (i.e., maximum sequence length).

\begin{table*}

\centering{

\centering
\resizebox{\textwidth}{!}{%
\begin{tabular}{@{}lcllllllllllr@{}}
\toprule
Sequence Per GPU & \multicolumn{3}{c}{4K}                                   & \multicolumn{3}{c}{8K}                                                  & \multicolumn{3}{c}{16K}                                                 & \multicolumn{3}{c}{32K}                                                 \\ \midrule
GPUs             & 8             & \multicolumn{1}{c}{16} & 64              & \multicolumn{1}{c}{8} & \multicolumn{1}{c}{16} & \multicolumn{1}{c}{64} & \multicolumn{1}{c}{8} & \multicolumn{1}{c}{16} & \multicolumn{1}{c}{64} & \multicolumn{1}{c}{8} & \multicolumn{1}{c}{16} & \multicolumn{1}{c}{64} \\ \midrule
Ulysses        & 14.6          & 29.31                  & OOM             & 49.74                 & 100.42                 & OOM                    & 176.23                & 349.85                 & OOM                    & 664.33                & 1367.85                & OOM                    \\
DistFlashAttn         & 20.46         & 54.58                  & 282.82          & 60.57                 & 136.99                 & 613.12                 & 205.29                & 436.65                 & 1788.33                & 763.13                & 1545.35                & 6427.15                \\
HSAP      & \textbf{13.4} & \textbf{26.87}         & \textbf{116.68} & \textbf{45.88}        & \textbf{92.54}         & \textbf{436.45}        & \textbf{168.44}       & \textbf{335.71}        & \textbf{1645.43}       & \textbf{662.98}       & \textbf{1350.28}       & \textbf{6156.32}       \\ \bottomrule
\end{tabular}%
}

}

\caption{\label{tbl-exp1}The speed of HSAP, DistFlashAttn, and Deepspeed-Ulysses on different sequence lengths. The speed is measured by milliseconds. OOM indicates that the model runs out of memory. The best performance is highlighted in bold.}

\end{table*}%

As illustrated in Table~\ref{tbl-exp1}, The experiment results present
that Deepspeed-Ulysess is highly sensitive to the number of attention
heads and cannot adapt to scenarios where the parallelism exceeds the
number of attention heads. Additionally, we can observe that with few
devices, HSAP primarily maintains a constant communication volume
through low-level intraSP, achieving slightly better performance than
Deepspeed-Ulysess under improved communication and memory management
strategies. As the number of devices increases, sinece HSAP uses two
orthogonal parallel groups to address the load imbalance issue, it
consistently outperforms the standalone Deepspeed-Ulysses and
DisFlashAttn methods in terms of performance.

\subsection{End-to-end training
Evaluation}\label{end-to-end-training-evaluation}

In practical training scenarios, we compared our algorithm with two
existing training methods to detect performance differences in
end-to-end scenario training. We selected FineWeb
\citep{lozhkov2024fineweb-edu},
Slim-Pajama\citep{together2023redpajama}, and
UltraChat\citep{ding2023enhancing}, as we believe these cover a wide
range of scenarios from pre-training to fine-tuning. The length
distribution of datasets used in these scenarios varies slightly,
resulting in different training performances, but our method
consistently outperforms the others Table~\ref{tbl-hsap}.

\begin{table*}

\centering{

\centering
\resizebox{\textwidth}{!}{%
\begin{tabular}{llllllll}
\hline
\multicolumn{2}{c}{Method} & \multicolumn{6}{c}{\textbf{HSAP/Packing/Batching}} \\ \hline
\multicolumn{2}{c}{Metric} & \multicolumn{3}{c|}{\textbf{throughput ↑}} & \multicolumn{3}{c}{\textbf{iter/s ↑}} \\ \hline
\multicolumn{2}{c|}{dataset} & \multicolumn{1}{c}{\textbf{fineweb}} & \multicolumn{1}{c}{\textbf{pajama}} & \multicolumn{1}{c|}{\textbf{ultrachat}} & \multicolumn{1}{c}{\textbf{fineweb}} & \multicolumn{1}{c}{\textbf{pajama}} & \multicolumn{1}{c}{\textbf{ultrachat}} \\ \hline
\multirow{4}{*}{16K} & \multicolumn{1}{l|}{1x8} & \textbf{2002.1}/1998.31/1997.5 & \textbf{2693.44}/2005.0/2003.36 & \multicolumn{1}{l|}{\textbf{2709.42}/1765.25/1766.04} & \textbf{1.09}/1.08/1.08 & \textbf{1.46}/1.09/1.09 & \textbf{1.67}/1.09/1.09 \\
 & \multicolumn{1}{l|}{2x4} & \textbf{2435.17}/2430.87/2442.83 & \textbf{2833.45}/2434.12/2444.45 & \multicolumn{1}{l|}{\textbf{2859.2}/2166.41/2162.06} & \textbf{1.32}/1.32/1.33 & \textbf{1.54}/1.32/1.33 & \textbf{1.76}/1.33/1.33 \\
 & \multicolumn{1}{l|}{4x2} & 2480.91/\textbf{2485.24}/2475.36 & \textbf{3064.3}/2484.06/2494.58 & \multicolumn{1}{l|}{\textbf{2901.98}/2191.22/2196.04} & \textbf{1.35}/1.35/1.34 & \textbf{1.66}/1.35/1.35 & \textbf{1.79}/1.35/1.35 \\
 & \multicolumn{1}{l|}{8x1} & 2863.02/\textbf{2883.24}/2877.61 & \textbf{3255.92}/2854.79/2875.19 & \multicolumn{1}{l|}{\textbf{3008.3}/2530.93/2541.83} & \textbf{1.55}/1.56/1.56 & \textbf{1.77}/1.55/1.56 & \textbf{1.85}/1.56/1.56 \\ \hline
\multirow{4}{*}{32K} & \multicolumn{1}{l|}{1x8} & \textbf{3642.88}/2067.07/1037.41 & \textbf{3235.0}/2070.18/2073.62 & \multicolumn{1}{l|}{\textbf{2129.07}/1200.87/512.46} & \textbf{0.99}/0.56/0.56 & \textbf{0.88}/0.56/0.56 & \textbf{0.99}/0.56/0.56 \\
 & \multicolumn{1}{l|}{2x4} & \textbf{3510.77}/2121.79/1070.83 & \textbf{3196.2}/2135.73/2123.38 & \multicolumn{1}{l|}{\textbf{2023.12}/1254.6/539.04} & \textbf{0.95}/0.58/0.58 & \textbf{0.87}/0.58/0.58 & \textbf{0.94}/0.59/0.59 \\
 & \multicolumn{1}{l|}{4x2} & \textbf{3574.18}/2060.45/1040.75 & \textbf{3286.72}/2057.39/2049.6 & \multicolumn{1}{l|}{\textbf{2125.67}/1213.83/521.49} & \textbf{0.97}/0.56/0.56 & \textbf{0.89}/0.56/0.56 & \textbf{0.99}/0.57/0.57 \\
 & \multicolumn{1}{l|}{8x1} & \textbf{3598.57}/2519.64/1277.91 & \textbf{3477.07}/2522.72/2511.99 & \multicolumn{1}{l|}{\textbf{2149.12}/1472.33/629.15} & \textbf{0.98}/0.68/0.69 & \textbf{0.94}/0.68/0.68 & \textbf{1.0}/0.69/0.68 \\ \hline
\multirow{4}{*}{64K} & \multicolumn{1}{l|}{1x8} & \textbf{3100.68}/1622.0/97.67 & \textbf{3414.77}/1618.27/1091.89 & \multicolumn{1}{l|}{\textbf{2994.03}/1251.3/593.16} & \textbf{0.42}/0.22/0.22 & \textbf{0.46}/0.22/0.22 & \textbf{0.52}/0.22/0.22 \\
 & \multicolumn{1}{l|}{2x4} & \textbf{3267.28}/1611.48/96.83 & \textbf{3281.22}/1614.85/1082.69 & \multicolumn{1}{l|}{\textbf{2811.5}/1254.9/589.35} & \textbf{0.44}/0.22/0.22 & \textbf{0.45}/0.22/0.22 & \textbf{0.49}/0.22/0.22 \\
 & \multicolumn{1}{l|}{4x2} & \textbf{3250.4}/1482.95/88.79 & \textbf{3312.26}/1438.61/970.22 & \multicolumn{1}{l|}{\textbf{2819.46}/1150.47/543.82} & \textbf{0.44}/0.2/0.2 & \textbf{0.45}/0.2/0.2 & \textbf{0.49}/0.2/0.2 \\
 & \multicolumn{1}{l|}{8x1} & \textbf{3459.34}/1900.28/114.41 & \textbf{3487.24}/1851.67/1243.12 & \multicolumn{1}{l|}{\textbf{2899.35}/1444.84/678.12} & \textbf{0.47}/0.26/0.26 & \textbf{0.47}/0.25/0.25 & \textbf{0.51}/0.25/0.25 \\ \hline
\end{tabular}%
}

}

\caption{\label{tbl-hsap} We validated the end-to-end performance of our method on Finewerb, pajama, and ultrachat, with throughput measured in tokens per second and iter/s representing the training speed of sequences with context length. The topology (intraSP×interSP) indicates how we mix sequence parallelism, where 1×8 and 8×1 are the cases where we degenerate to DistFlashAttn and Ulysess, respectively.}

\end{table*}%

In Table~\ref{tbl-hsap}, we used two pre-training datasets (fineweb and
pajama) and one SFT dataset (ultra-chata) to validate the superiority of
HSAP in addressing the long-tail problem under various N×M topologies (N
representing interSP and M representing intraSP). Compared to pure
batching and packing methods, HSAP offers higher throughput and training
efficiency. As the sequence length increases, the disadvantage of the
batching method is gradually apparent due to the requirement to pad all
sequences to a uniform length. In contrast to the packing method, the
sequence-aware parallel strategies better leverages the overlap between
communication and computation.

\subsection{Sequence Scalability
Comparison}\label{sequence-scalability-comparison}

\begin{figure}

\centering{

\includegraphics{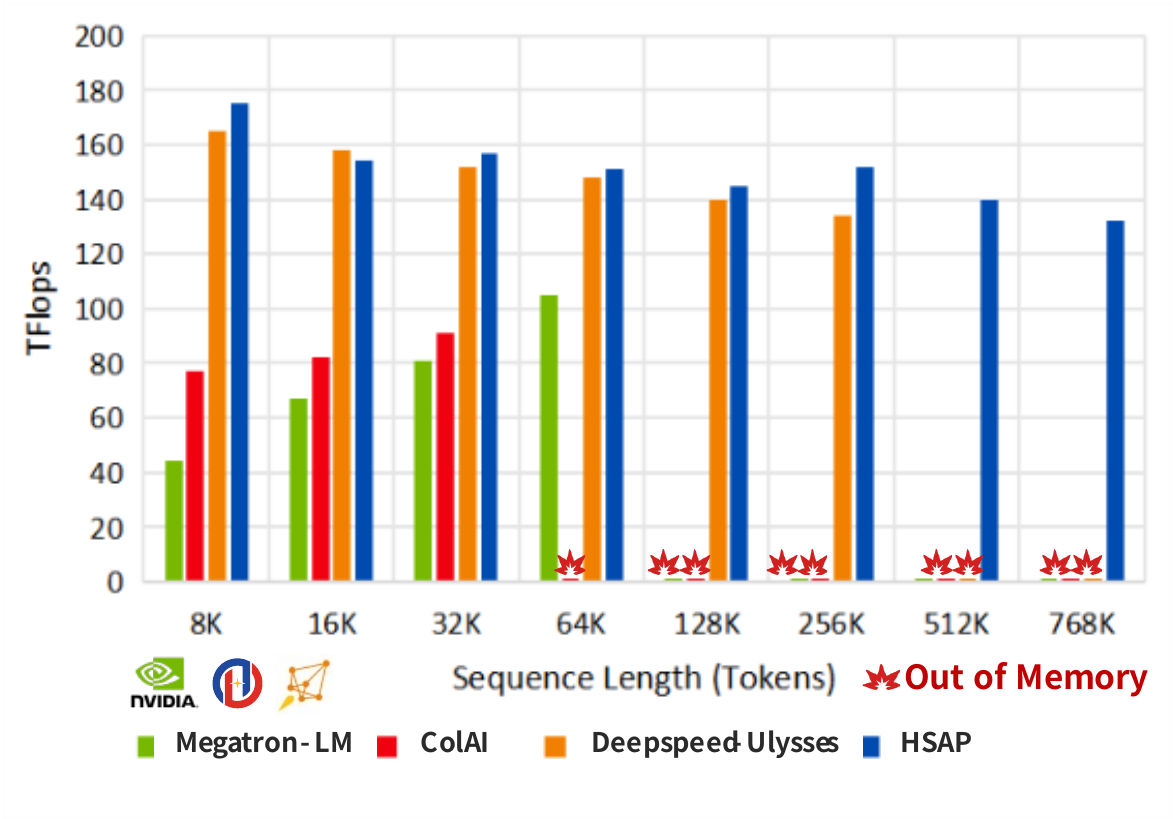}

}

\caption{\label{fig-cost}Evaluation Megatron vs ColAL-SP vs Ulysses and
HSAP on GPT-30B parameter model with dense attention (64 GPUs).}

\end{figure}%

We compare Ulysses, HSAP, Megatron-LM, and ColAI-SP for 30B GPT models
running various sequence lengths. We chose the sequence parallelism
degree and micro-batch size that produced the best performance (measured
as TFLOPs) for the four methods, this we call optimal (batch
size-sequence length) configurations. For HSAP, we utilize the
integrated benefits of both inter-group and intra-group sequence
parallelism, ensuring efficient handling of extremely long sequences
while maintaining high performance. As illustrate in
Figure~\ref{fig-cost}, HSAP can accommodate ultra-long sequences without
adding extra burden.

\section{Conclusion}\label{conclusion}

In conclusion, this paper presents an innovative approach to addressing
the challenges of sequence parallelism on hybrid-context long packed
sequences for training large language models (LLMs). By introducing the
Sequence-Aware Parallelism (SAP) algorithm, we leverage the just-in-time
(JIT) compilation idea to manage the intricate scheduling and
organization of incomplete QKV tensor transmission and irregular
attention patterns computation. Our approach not only ensures the
efficient and harmonious arrangement of sophisticated P2P communications
but also integrates the benefits of both inter-group and intra-group
sequence parallelism through the Hierarchical Sequence-Aware Parallelism
(HSAP) framework. This integration enhances memory and communication
management, providing a scalable solution for handling extremely long
sequences.

Our comprehensive experiments demonstrate the superiority of our
proposed SAP algorithm and HSAP framework over existing sequence
parallelism approaches. We successfully extend sequence lengths to over
512K, achieving the highest efficiency in terms of tokens per GPU second
(TGS) when training on real hybrid-context datasets. Additionally, our
method shows faster attention computation on packed sequences compared
to current state-of-the-art sequence parallelism paradigms. These
results underscore the significant advancements our approach offers in
the realm of sequence parallelism for LLM training. The primary
limitation of our paper is that due to lack of enough computational
resourses, we did not go through the complete pretraining procedure of
large scaled models to examine its performance. We hope to supplement
this experiment in the furture.

\section{Limitations}\label{limitations}

The primary limitation of our paper is that due to lack of enough
computational resourses in a period of time, we did not go through the
complete pretraining procedure of large scaled models to examine its
performance. We hope to supplement this experiment in the furture.

We also think of a promising direction to leverage Triton or PyTorch
CUDA C++ extensions to fuse the copy and rescale operations directly
into the attention computation. This fusion can potentially enhance
computational efficiency by reducing the overhead associated with
separate operations and streamlining the data flow within the GPU.
Additionally, integrating these operations at a lower level within the
CUDA framework allows for more fine-grained optimization and better
utilization of the GPU's capabilities. By exploring these advanced
techniques, we can further improve the performance of our method.

\renewcommand{\bibsection}{}
\bibliography{custom.bib}

\appendix

\end{document}